# Risk Agoras: Dialectical Argumentation for Scientific Reasoning

**Peter McBurney** and **Simon Parsons**
Department of Computer Science
University of Liverpool
Liverpool L69 7ZF United Kingdom
P.J.McBurney,S.D.Parsons@csc.liv.ac.uk

## Abstract

We propose a formal framework for intelligent systems which can reason about scientific domains, in particular about the carcinogenicity of chemicals, and we study its properties. Our framework is grounded in a philosophy of scientific enquiry and discourse, and uses a model of dialectical argumentation. The formalism enables representation of scientific uncertainty and conflict in a manner suitable for qualitative reasoning about the domain.

## 1 INTRODUCTION

We seek to build intelligent systems which can reason autonomously about the risk of carcinogenicity of chemicals, drawing on whatever theoretical or experimental evidence is available. In earlier work (McBurney & Parsons 1999), reviewing the literature on methods of carcinogen risk assessment, we listed the different types of evidence adduced to support these claims, which may be in the form of: experimental results on tissue cultures, animals or human epidemiological studies; analytical comparisons with known carcinogens; or explication of biomedical causal pathways. Evidence from these different sources may conflict, and carcinogen risk assessment usually involves the comparison and resolution of multiple evidence (E.P.A. U.S.A. 1986; Graham, Green, & Roberts 1988). In representing this domain, it therefore seems appropriate to use some form of argumentation (so that the reasons for claims can be represented in association with the claims themselves), and within a dialectical framework (so that cases for and against a particular claim can be compared). In particular, dialectical argumentation enables the representation of uncertainty in the underlying scientific knowledge base. This paper presents such a dialectical formalism for an intelligent system, which we termed a Risk Agora in our earlier work. We begin by examining the nature of scientific discourse.

## 2 SCIENTIFIC DISCOURSE

### 2.1 A MODEL OF SCIENTIFIC ENQUIRY

Our chosen application domain is a scientific one. To represent this domain, therefore, we seek to ground our formalism in a philosophical model of scientific enquiry. Firstly, we require a theory of the nature of modern science. Following Pera (1994), we view the enterprise of science as a three-person dialogue, involving a scientific investigator, Nature and a skeptical scientific community. In Pera's model, the investigator proposes theoretical explanations of scientific phenomena and undertakes scientific experiments to test these. The experiments lead to "replies" from Nature in the form of experimental evidence. However, Nature's responses are not given directly or in a pure form, but are mediated through the third participant, the scientific community, which interprets the evidence, undertakes a debate as to its meaning and implications, and eventually decides in favor or against proposed theoretical explanations. The consequence of this model for our formalism is that we provide Nature with a formal role, but manifest it through those of the other participants.

But Pera's model of modern science as a dialogue game could apply to many other human dialogues, most of which do not share science's success in explaining and predicting natural phenomena. Our model of science therefore requires an explanation of its success. Some philosophers of science believe this is due to the application of universal principles of assessment of proposed scientific theories, such as the falsificationism of Popper or the confirmationism of Carnap. However, we do not share these views, instead believing, with Feyerabend (1993), that the standards of assessment used by any scientific community are domain-, context- and time-dependent. This view, that there are neither universal nor objective standards by which scientific theories can be judged, was called "epistemological anarchism" by Lakatos (Lakatos & Feyerabend 1999). Instead of universal principles of assessment of theories, we believe science's success arises in part from applying two normative principles of conduct: firstly, that every the-oretical explanation proposed by a scientific investigator is contestable by anyone; and secondly, that every theoretical explanation adopted by a scientific community is defeasible. In other words, all scientific theories, no matter how compelling, are always tentative, being held only until better explanations are found, and anyone may propose these.[1]

To build an intelligent system based on these principles, we therefore require a (normative) model of scientific discourse which enables contestation and defeasibility of claims. Our model has several components. At the highest level, we are attempting to model a discourse between reasonable, consenting scientists, who accept or reject arguments only on the basis of their relative force. An influential model for debates of this type is the philosophy of Discourse Ethics developed by Habermas (1991) for debates in ethical and moral domains. Our formalism therefore draws on Habermas, in particular his rules of discourse first fully articulated by Alexy (1990), and these form the basis of the desired properties of the Agora formalism presented later in this section.[2]

Next, within this structure, we wish to be able to model dialogues in which different participants variously posit, assert, contest, justify, qualify and retract claims. To represent such activity requires a model of an argument, and we use Toulmin's (1958) model, within a dialectical framework. To embody our belief in epistemological anarchism, we permit participants to contest any component of a scientific argument: its premises; its rules of inference (Toulmin's "warrants"); its degrees of support (his "modalities"); and its consequences. We believe this is exactly what real scientists do when confronted with new theoretical explanations of natural phenomena (Feyerabend 1993). When a scientific claim is thus contested, its proponent may respond, not only by retracting it, but by qualifying it in some way, perhaps reducing its scope of applicability. Naess (1966) called this process "precizating", and we seek to enable such responses in the system. We thus ground our formalism for the Agora in a model of scientific discourse as dialectical argumentation.[3]

### 2.2 DESIRED AGORA PROPERTIES

As mentioned, we desire our Agora formalism to satisfy the rules for a reasoned discourse proposed by Alexy (1990), which are listed here. In restating these, we have modified and re-ordered them slightly, and have ignored rules which deal specifically with discussion of ethical matters. Also, because our formalism is intended for debate regarding only one chemical at a time, we have ignored Alexy's rules regarding the relevance of utterances. We have also added a property concerning precization.

**P1** Anyone may participate in the Agora, and they may execute dialogue moves at any time, subject only to move-specific conditions (defined below).

**P2** Participation entails acceptance of the semantics for the logical language used, and of the associated modality (degrees of support) dictionaries.

**P3** Any participant may assert any claim or consequence of a claim, but may do so only when they have a grounded argument for the claim (respectively, a consequential argument from the claim).

**P4** Any participant may question or challenge any claim or any consequence of a claim.

**P5** Any participant who asserts a claim (respectively, a consequence of a claim) must provide a valued grounded argument for that claim (respectively, a valued consequential argument from the claim) if queried or challenged by another participant.

**P6** Any participant may question or challenge the grounds, the rules of inference or the modalities for any claim.

**P7** Whenever a participant asserts a valued grounded argument for a claim (or a valued consequential argument from a claim), any other participant may assert a valued grounded argument (respectively, a valued consequential argument) for the same claim with different dictionary values.

**P8** A participant who has provided a grounded argument for a claim which has been challenged should be able to respond by qualifying (precizating) the original claim or argument.

**P9** Any participant who provides a grounded argument for, or a consequential argument from, a claim is not required to provide further defence if no counter-arguments are provided by other participants.

**P10** No participant may contradict him or herself.

## 3 THE RISK AGORA FORMALISM

### 3.1 PRELIMINARY DEFINITIONS

We begin by assuming the system is intended to represent debate regarding the carcinogenicity of a specific chemical, and that statements concerning this can be expressed in a propositional language $\mathcal{L}$, whose well-formed formulae (wffs) we denote by lower-case Greek letters. Subsets of $\mathcal{L}$ (i.e. sets of wffs) are denoted by upper-case Greek letters, and $\mathcal{L}$ is assumed closed under the usual connectives. We assume multiple modes of inference (warrants) are possible, these being denoted by $\vdash_i$. These may include non-deductive modes of reasoning, and we make no presumptions regarding their validity in any truth model. We assume a finite set of debate participants, denoted by $\mathcal{P}_i$, who are permitted to introduce new wffs and new modes of inference at any time. We denote Nature, in its role in the debate, by $\mathcal{P}_N$.

---

[1] These two principles are each necessary to explain science's success, but not sufficient.

[2] Alexy's rules have some similarity with Grice's (1975) Maxims for Conversation.

[3] Further details of our philosophy of science are contained in (McBurney & Parsons 2000b).

**Definition 1:** *A grounded argument for a claim $\theta$, denoted $\mathcal{A}(\rightarrow \theta)$, is a 3-tuple $(G, R, \theta)$, where $G = (\Theta_0, \theta_1, \Theta_1, \theta_2, \ldots, \Theta_{n-2}, \theta_{n-1}, \Theta_{n-1})$ is an ordered sequence of wffs $\theta_i$ and possibly-empty sets of wffs $\Theta_i$, with $n \geq 1$ and with $R = (\vdash_1, \vdash_2, \ldots, \vdash_n)$ an ordered sequence of inference rules such that:*

$$\Theta_0 \vdash_1 \theta_1,$$

$$\theta_1, \Theta_1 \vdash_2 \theta_2,$$

$$\vdots$$

$$\theta_{n-1}, \Theta_{n-1} \vdash_n \theta.$$

In other words, each $\theta_k$ ($k = 1, \ldots, n-1$) is derived from the preceding wff $\theta_{k-1}$ and set of wffs $\Theta_{k-1}$ as a result of the application of the k-th rule of inference, $\vdash_k$. The rules of inference in any argument may be non-distinct. We call the set $\{\theta_{k-1}\} \cup \Theta_{k-1}$ the *grounds* (or *premises*) for $\theta_k$.

**Definition 2:** *A consequential argument from a claim $\theta$, denoted $\mathcal{A}(\theta \rightarrow)$, is a 3-tuple $(\theta, R, C)$, where $C = (\Theta_0, \theta_1, \Theta_1, \theta_2, \ldots, \Theta_{n-2}, \theta_{n-1}, \Theta_{n-1}, \theta_n)$ is an ordered sequence of wffs $\theta_i$ and possibly-empty sets of wffs $\Theta_i$, with $n \geq 1$, and with $R = (\vdash_1, \vdash_2, \ldots, \vdash_n)$ an ordered sequence of inference rules such that:*

$$\theta, \Theta_0 \vdash_1 \theta_1,$$

$$\theta_1, \Theta_1 \vdash_2 \theta_2,$$

$$\vdots$$

$$\theta_{n-1}, \Theta_{n-1} \vdash_n \theta_n.$$

In other words, the wffs $\theta_k$ in $C$ are derivations from $\theta$ arising from the successive application of the rules of inference in $R$, and we call each $\theta_k$ in $C$ a *consequence* of $\theta$.

In order that participants may effectively state and contest degrees of commitment to claims, we require a common dictionary of degrees of commitment or support (what Toulmin called "modalities"). Our formalism will support any agreed dictionary, whether quantitative (such as a set of probability values or belief measures) or qualitative (such as non-numeric symbols or linguistic qualifiers), provided there is a partial order on its elements. We define dictionaries for modalities for claims, grounds, consequences and rules of inference.

**Definition 3:** *Four modality dictionaries are defined as follows, each being a (possibly infinite) set of elements having a partial order. The claims dictionary is denoted by $\mathcal{D}_C$, the grounds dictionary by $\mathcal{D}_G$, the consequences dictionary by $\mathcal{D}_Q$, and the inference dictionary by $\mathcal{D}_I$.*

Because claims, grounds and consequences are all elements of the same language $\mathcal{L}$, two or more of the dictionaries $\mathcal{D}_C$, $\mathcal{D}_G$ and $\mathcal{D}_Q$ may be the same. However, a distinct dictionary will generally be required for $\mathcal{D}_I$.[4] Because of our belief in epistemological anarchism, we do not specify rules of assignment of dictionary labels by participants in the Agora. In particular, the labels assigned to the conclusions and consequences of arguments are not constrained by those assigned to premises or rules of inference.

**Example 1:** *The generic argumentation dictionary defined for assessment of risk by (Krause et al. 1998) is an example of a linguistic dictionary for statements about claims, grounds or consequences, comprising the set: {Certain, Confirmed, Probable, Plausible, Supported, Open}. The elements of this dictionary are listed in descending order, with each successive label indicating a weaker belief in the claim.*

**Example 2:** *Two examples of Inference Dictionaries are $\mathcal{D}_I = \{\text{Valid, Invalid}\}$ and $\mathcal{D}_I = \{\text{Acceptable, Sometimes Acceptable, Open, Not Acceptable}\}$.*

**Definition 4:** *A valued grounded argument for a claim $\theta$, denoted $\mathcal{A}(\rightarrow \theta, D)$, is a 4-tuple $(G, R, \theta, D)$, where $(G, R, \theta)$ is a grounded argument for $\theta$ and $D = (\tilde{d}_0, \tilde{d}_1, \ldots, \tilde{d}_{n-1}, d_\theta, r_1, r_2, \ldots, r_n)$ is an ordered sequence of labels and vectors of labels, with each $\tilde{d}_i$ a vector of dictionary labels from $\mathcal{D}_C$ (for $i = 0, \ldots, n-1$), with $d_\theta \in \mathcal{D}_C$ and with $r_i \in \mathcal{D}_I$ (for $i = 1, \ldots, n$). Each vector $\tilde{d}_i$ comprises those values of the Claims Dictionary assigned to grounds $\{\theta_i\} \cup \Theta_i$, the element $d_\theta$ is that value of the Claims Dictionary assigned to $\theta$ and each element $r_i$ is that value of the Inference Dictionary assigned to $\vdash_i$. A valued consequential argument from a claim $\theta$, denoted $\mathcal{A}(\theta \rightarrow, D)$, is defined similarly.*

### 3.2 DISCOURSE RULES

We next define the rules for discourse participants, building on the definitions above. Moves are denoted by 2-ary or 3-ary functions of the form *name($\mathcal{P}_i$: . )*, where the first argument denotes the participant executing the move. If the move responds to an earlier move by another participant, that earlier move is the second argument. Arguments are separated by colons. In Section 4, we will show that these rules give operational effect to the Desired Properties.

---

[4] In (McBurney & Parsons 2000c), we model degrees of acceptability of inference rules.

**Rule 1: Query and Assertion Moves**

**1.1 Pose Claim:** Any participant $\mathcal{P}_i$ at any time may move:
$$pose(\mathcal{P}_i :\rightarrow \theta?)$$
which asks the Agora if there is a grounded argument for $\theta$.

**1.2 Propose Claim:** Any participant $\mathcal{P}_i$ at any time may propose a claim with move:
$$propose(\mathcal{P}_i : (\theta, d_\theta))$$
where $\theta \in \mathcal{L}$ and $d_\theta \in \mathcal{D}_C$, which informs the Agora that $\mathcal{P}_i$ has a valued grounded argument for $\theta$, and has assigned it a modality of $d_\theta$.

**1.3 Assert Claim:** Any participant $\mathcal{P}_i$ at any time may assert a claim with move:
$$assert(\mathcal{P}_i : (\theta, d_\theta))$$
where $\theta$ is a wff and $d_\theta \in \mathcal{D}_C$, which informs the Agora that $\mathcal{P}_i$ has a valued grounded argument for $\theta$, which she believes is compelling.

**1.4 Query Claim:** Whenever a *propose* or *assert* move relating to $(\theta, d_\theta)$ has been made by participant $\mathcal{P}_i$, any other participant $\mathcal{P}_j$ may move:
$$query(\mathcal{P}_j :propose(\mathcal{P}_i : (\theta, d_\theta)))$$
or
$$query(\mathcal{P}_j :assert(\mathcal{P}_i : (\theta, d_\theta))).$$
These ask participant $\mathcal{P}_i$ to provide her valued grounded argument for $\theta$, which she must provide immediately with move:
$$show\_arg(\mathcal{P}_i : \mathcal{A}(\rightarrow \theta, D)).$$

**1.5 Show Grounded Argument:** Any participant $\mathcal{P}_i$ may at any time provide a valued grounded argument for $\theta$ with the move:
$$show\_arg(\mathcal{P}_i : \mathcal{A}(\rightarrow \theta, D)).$$

**1.6 Pose Consequence:** A participant $\mathcal{P}_k$ may at any time move:
$$pose\_cons(\mathcal{P}_k : \theta \rightarrow ?)$$
which asks the Agora if there is a consequential argument from $\theta$.

**1.7 Propose Consequence:** Similarly to *Propose Claim*, a participant may move:
$$propose\_cons(\mathcal{P}_i : (\theta, \phi, d_\phi))$$
where $\phi$ is a consequence of $\theta$.

**1.8 Assert Consequence:** Similarly to *Assert Claim*, a participant may move:
$$assert\_cons(\mathcal{P}_i : (\theta, \phi, d_\phi))$$
where $\phi$ is a consequence of $\theta$.

**1.9 Query Consequence:** Similarly to *Query Claim*, a participant may move:
$$query\_cons(\mathcal{P}_j :propose(\mathcal{P}_i : (\theta, \phi, d_\phi))).$$

**1.10 Show Consequential Argument:** Any participant $\mathcal{P}_i$ may at any time provide a valued consequential argument from $\theta$ with the move:
$$show\_cons(\mathcal{P}_i : \mathcal{A}(\theta \rightarrow, D)).$$

**1.11 Propose Mode of Inference:** Any participant $\mathcal{P}_i$ at any time may move:
$$propose\_inf(\mathcal{P}_i : (\vdash_t, r_t))$$
where $\vdash_t$ is a mode of inference and $r_t \in \mathcal{D}_I$. This move informs the community that participant $\mathcal{P}_i$ believes that $\vdash_t$ is a mode of inference of strength at least $r_t$.

Note that the query and assertions rules are not symmetric between grounded and consequential arguments; participants may only propose or assert claims for which they have grounded arguments, but they need not necessarily have considered the consequences of these claims. Next, we explicitly define the *Contest Claim* rule, with other contestation rules being defined similarly. For brevity in the following, we sometimes write $\mathcal{A}$ for $\mathcal{A}(\rightarrow \theta, D)$.

**Rule 2: Contestation Moves**

**2.1 Contest Claim:** Whenever *propose* or *assert* relating to $(\theta, d_\theta)$ has been moved by participant $\mathcal{P}_i$, any other participant $\mathcal{P}_j$ may contest this by moving:
$$contest(\mathcal{P}_j :propose(\mathcal{P}_i : (\theta, d_\theta)))$$
or
$$contest(\mathcal{P}_j :assert(\mathcal{P}_i : (\theta, d_\theta))).$$
If any participant $\mathcal{P}_k$ subsequently queries this contestation with:
$$query(\mathcal{P}_k :contest(\mathcal{P}_j :propose(\mathcal{P}_i : (\theta, d_\theta))))$$
(or likewise for *assert*), participant $\mathcal{P}_j$ must respond immediately, either with an assignment of an alternative modality $d'_\theta$ for claim $\theta$, thus:
$$propose(\mathcal{P}_j : (\theta, d'_\theta))$$

or
$$assert(\mathcal{P}_j : (\theta, d'_\theta))$$
(where $d'_\theta \neq d_\theta$), or with a stronger assertion of the negation of $\theta$, thus:
$$propose(\mathcal{P}_j : (\neg\theta, d'_\theta))$$
or
$$assert(\mathcal{P}_j : (\neg\theta, d'_\theta))$$
(where $d'_\theta > d_\theta$).

### 2.2 Contest Ground:
$$contest\_ground(\mathcal{P}_j : show\_arg(\mathcal{P}_i : \mathcal{A} : (\theta_t, d_{\theta_t}))).$$

### 2.3 Contest Inference:
$$contest\_inf(\mathcal{P}_j : show\_arg(\mathcal{P}_i : \mathcal{A} : \vdash_t)).$$

### 2.4 Contest Modality:
$$contest\_mod(\mathcal{P}_j : show\_arg(\mathcal{P}_i : \mathcal{A}(\to \theta, D))).$$

### 2.5 Contest Consequence:
$$contest\_cons(\mathcal{P}_j : show\_cons(\mathcal{P}_i : \mathcal{A} : (\theta_t, d_{\theta_t}))).$$

### Rule 3: Participant Resolution Moves

### 3.1 Accept Proposed Claim:
Whenever a claim has been proposed by $\mathcal{P}_i$ and its grounds demonstrated by moving:
$$show\_arg(\mathcal{P}_i : \mathcal{A}(\to \theta, D)),$$
any other participant $\mathcal{P}_j$ may declare that they accept the proposed claim, with move:
$$accept\_prop(\mathcal{P}_j : show\_arg(\mathcal{P}_i : \mathcal{A}(\to \theta, D))).$$
This move is identical with the sequence:
$$propose(\mathcal{P}_j : (\theta, d_\theta))$$
$$show\_arg(\mathcal{P}_j : \mathcal{A}(\to \theta, D)).$$

### 3.2 Accept Asserted Claim: Similarly to *accept_prop*:
$$accept\_assert(\mathcal{P}_j : show\_arg(\mathcal{P}_i : \mathcal{A}(\to \theta, D))).$$

### 3.3 Change Modalities:
Any participant $\mathcal{P}_i$ who proposes or asserts a claim for $\theta$, and follows this with a demonstration of a valued grounded argument for $\theta$ by moving:
$$show\_arg(\mathcal{P}_i : \mathcal{A}(\to \theta, D)),$$
may subsequently revise her assignment of modalities with a later move of:
$$show\_arg(\mathcal{P}_i : \mathcal{A}(\to \theta, D')),$$
where $D' \neq D$. Likewise, declarations of modal beliefs expressed in other moves (e.g. in *accept_assert*) may also be revised by subsequently executing the same move with a different set of dictionary values.

### 3.4 Accept Mode of Inference: Similarly to *accept_prop*:
$$accept\_inf(\mathcal{P}_j : propose\_inf(\mathcal{P}_i : (\vdash_t, r_t))).$$

### 3.5 Accept Consequence: Similarly to *accept_prop*:
$$accept\_cons(\mathcal{P}_j : show\_cons(\mathcal{P}_i : \mathcal{A}(\theta \to, D))).$$

### 3.6 Precizate Claim:
Any participant $\mathcal{P}_i$ who proposes or asserts a claim for $\theta$, and follows this with a demonstration of a valued grounded argument for $\theta$ by:
$$show\_arg(\mathcal{P}_i : \mathcal{A}(\to \theta, D))$$
may subsequently qualify her argument with:
$$prec(\mathcal{P}_i : show\_arg(\mathcal{P}_i : \mathcal{A}(\to \theta, D)): \mathcal{A}'(\to \theta, D'))$$
where $\mathcal{A}'(\to \theta, D')$ is an argument for $\theta$ identical with $\mathcal{A}(\to \theta, D)$ except that: (a) it begins from ground $\Phi \cup \Theta_0$ instead of $\theta_0$, where $\Phi$ is not equal to $\{\theta\}$ nor to any ground of $\theta$, and (b) $D'$ may be different to $D$.

### 3.7 Retract Claim: Any participant $\mathcal{P}_i$ who asserts:
$$assert(\mathcal{P}_i : (\theta, d_\theta))$$
may at any time subsequently withdraw the claim by:
$$retract(\mathcal{P}_i : assert(\mathcal{P}_i : (\theta, d_\theta))).$$
Likewise, for those claims by others accepted by $\mathcal{P}_i$.

### 3.8 No contradiction:
Any participant $\mathcal{P}_i$ who asserts (or accepts an assertion for) $\theta$ may not at any time subsequently assert (or accept an assertion for) $\neg\theta$, unless they have in the interim moved:
$$retract(\mathcal{P}_i : assert(\mathcal{P}_i : (\theta, d_\theta)))$$
(or, respectively, its equivalent for accepted claims).

## 3.3 DIALOGUE RULES

**Definition 5:** *A* Dialogue *is a finite sequence of discourse moves by participants in the Agora, in accordance with the rules above.*

As in (Hamblin 1971; Walton & Krabbe 1995; Amgoud, Maudet, & Parsons 2000), we define sets called Commitment Stores which contain the proposals and assertions made by participants, both individually and for the Agora as a community, and track these as they change.

**Definition 6:** *The commitment store of player $\mathcal{P}_i$, $i = 1, 2, \ldots$, denoted $\mathcal{CS}(\mathcal{P}_i)$, is a possibly empty set $\{(\theta, d_\theta) \mid \theta \in \mathcal{L}, d_\theta \in \mathcal{D}_C\}$. Each $d_\theta$ is the claim dictionary value assigned by $\mathcal{P}_i$ to $\theta$.*

The values in participants' stores are updated by the following rule:

**Rule 4: Participant Commitment Store Update:** Whenever participant $\mathcal{P}_i$ executes the moves

$$propose(\mathcal{P}_i : (\theta, d_\theta)),$$

$$accept\_prop(\mathcal{P}_i : propose(\mathcal{P}_j : (\theta, d_\theta))),$$

$$assert(\mathcal{P}_i : (\theta, d_\theta)),$$

$$accept\_assert(\mathcal{P}_i : assert(\mathcal{P}_j : (\theta, d_\theta)))$$

or their equivalents, then the tuple $(\theta, d_\theta)$ is inserted into $\mathcal{CS}(\mathcal{P}_i)$. Whenever participant $\mathcal{P}_i$ executes a retraction move for $(\theta, d_\theta)$, the tuple $(\theta, d_\theta)$ is removed from $\mathcal{CS}(\mathcal{P}_i)$. Similarly, whenever $\mathcal{P}_i$ executes a Change Modality move for $(\theta, d_\theta)$, the value of $(\theta, d_\theta)$ in $\mathcal{CS}(\mathcal{P}_i)$ is revised.

We next define an analogous concept for Nature, with claims inserted into Nature's Commitment Store on the basis of the debate at that point in the Agora. This could be achieved in a number of ways. For example, a skeptical community could define Nature's modality for a claim $\theta$ to be the minimum claim modality assigned by any of those Participants claiming or supporting $\theta$. A credulous community could instead assign to Nature the maximum claim modality assigned by any of the participants to $\theta$. Variations on these approaches could utilize majority opinion or weighted voting schemes.

Because we wish to model dialectical discourse, we have instead chosen to assign Nature's modalities on the basis of the existence of arguments for and against the claim. To do this, we draw on the generic argumentation dictionary for debates about carcinogenicity of chemicals presented in (Krause *et al.* 1998), which is based on Toulmin's (1958) schema. We begin by defining certain relationships between arguments and then the Claims Dictionary for Nature.

**Definition 7:** *An argument $\mathcal{A}(\rightarrow \theta) = (G, R, \theta)$ is consistent if $G = (\Theta_0, \theta_1, \Theta_1, \theta_2, \ldots, \Theta_{n-2}, \theta_{n-1}, \Theta_{n-1})$ is consistent, that is if there do not exist $\alpha, \beta \in \Theta_0 \cup \{\theta_1\} \cup \Theta_1 \cup \{\theta_2\} \cup \ldots \cup \Theta_{n-1}$ such that $\neg\beta$ is a consequence of $\alpha$.*

**Definition 8:** *Let $\mathcal{A}(\rightarrow \theta) = (G, R, \theta)$ and $\mathcal{B}(\rightarrow \phi) = (H, S, \phi)$ be two arguments, where $G = (\Theta_0, \theta_1, \Theta_1, \theta_2, \ldots, \theta_{n-1}, \Theta_{n-1})$. We say that $\mathcal{B}(\rightarrow \phi)$ rebuts $\mathcal{A}(\rightarrow \theta)$ if $\phi \equiv \neg\theta$. We say that $\mathcal{B}(\rightarrow \phi)$ undercuts $\mathcal{A}(\rightarrow \theta)$ if, for some $\alpha \in \Theta_0 \cup \{\theta_1\} \cup \Theta_1 \cup \{\theta_2\} \cup \ldots \cup \Theta_{n-1}$, $\alpha \equiv \neg\phi$.*

**Definition 9:** *The claims dictionary for Nature is the set $\mathcal{D}_{C,N} = \{Certain, Confirmed, Probable, Plausible, Supported, Open\}$.*

**Definition 10:** *The commitment store of Nature, denoted $\mathcal{CS}(\mathcal{P}_N)$, is a non-empty set $\{(\theta, d_{\theta,N}) \mid \theta \in \mathcal{L}, d_{\theta,N} \in \mathcal{D}_{C,N}\}$. Each $d_{\theta,N}$ is the claim modality assigned by the Agora community on Nature's behalf to $\theta$, in accordance with the next two rules.*

**Rule 5: Nature's Modalities:** The modality $d_{\theta,N}$ of Nature for the claim $\theta$ is assigned as follows:

- If $\theta$ is a wff for which no grounded argument has yet been provided by a participant, then $d_{\theta,N}$ is assigned the value *Open*.

- If $\theta$ is a wff for which at least one grounded argument has been provided by a participant, then $d_{\theta,N}$ is assigned the value *Supported*.

- If $\theta$ is a wff for which a grounded and consistent argument has been provided by a participant, then $d_{\theta,N}$ is assigned the value *Plausible*.

- If $\theta$ is a wff for which a grounded and consistent argument has been provided by a participant, and for which no rebutting arguments have been provided, then $d_{\theta,N}$ is assigned the value *Probable*.

- If $\theta$ is a wff for which a grounded and consistent argument has been provided by a participant, and for which neither rebutting nor undercutting arguments have been provided by participants, then $d_{\theta,N}$ is assigned the value *Confirmed*.

- If $\theta$ is a logical tautology, then $d_{\theta,N}$ is assigned the value *Certain*.

**Rule 6: Nature Commitment Store Update:** The entries in $\mathcal{CS}(\mathcal{P}_N)$ are updated after each legal move by Agora participants.

### 3.4 ARCHITECTURE AND USER INTERFACE

We anticipate the Risk Agora system being used to represent a completed or on-going scientific debate, but not in real-time. Once instantiated with a specific knowledge base in this way, the Agora could be used for a number of different purposes, which led us (McBurney & Parsons 1999), to propose a layered architecture for the Agora, corresponding to these different functions. The main purposes to be fulfilled are: (a) automated reasoning to find arguments for, and the consequences of, particular claims; (b) comparison of the various arguments for and against a claim; and (c) development of an overall case for a claim, coherently combining all the arguments for and against it.

## 4 AGORA PROPERTIES

The rules defined in the previous section were intended to operationalize the desired Agora properties of Section 2.2. We now verify that this is indeed the case.

**Theorem 1:** *The Agora system defined in Section 3 has Properties P1 through P10.*
**Proof.** This is straightforward, from the definitions of the permitted moves. Thus, Properties P1 and P2 are fulfilled through the overall system design; Property P3 by Rules 1.1-1.3 and 1.6-1.8; Property P4 by Rules 1.4, 1.9, 2.1 and 2.5; Property P5 by Rules 1.2, 1.3, 1.5, 1.7, 1.8 and 1.10; Property P6 by Rules 2.2-2.4; Property P7 by Rules 3.1-3.5; Property P8 by Rule 3.6; Property P9 by Rules 1 and 2; and Property P10 by Rule 3.8. □

Moreover, we can use the definition of the claim modalities for Nature provided by Rule 5 to construct a valuation function on wffs and to define a notion of "proof" of claims, as follows.

**Definition 11:** Natural valuation *is a function $v_N$ defined from the set of wffs of $\mathcal{L}$ to the set $\{0,1\}$, such that $v_N(\theta) = 1$ precisely when $d_{\theta,N} = Confirmed$; otherwise, $v_N(\theta) = 0$.*

**Definition 12:** A *provisional proof for a claim $\theta$ is a grounded and consistent argument for $\theta$ for which neither rebuttal nor undercutting arguments exist.*

Our belief in the defeasibility of all scientific claims leads us to use the term "provisional proof" rather than "proof." Likewise, we can think of a natural valuation equal to 1 as signifying *"Currently Accepted as True"* (or *"Defeasibly True"*) and 0 as *"Not Currently Accepted as True."* Our definition of natural valuation thus says that a claim is defeasibly true iff there are no arguments attacking it. We could readily define additional valuation functions which capture degrees of conviction regarding the truth of claims, mapping, for instance, to *Probable* or to *Plausible*. With the definitions above, we can now prove soundness of provisional proofs in the Agora, with respect to the natural valuation function.

**Theorem 2:** *With the notion of provisional proof, the Agora is consistent and complete with respect to the Natural Valuation Function $v_N$, provided that all grounded arguments for claims are eventually asserted by some Participant.*
**Proof.** Consistency here says that all claims $\theta$ for which there exists a provisional proof are also assigned a valuation of 1 by the function $v_N$. Completeness says, conversely, that all claims $\theta$ which are assigned a valuation of 1 by $v_N$ also have a provisional proof. Both of these follow from our definitions of $v_N$ and of provisional proof, unless a consistent grounded argument for a claim $\theta$ exists but is not asserted by any Participant. □

The model of science we have adopted asserts that scientific claims are regarded as "defeasibly true" only when the relevant scientific community agrees to so regard them. (After all, even if a transcendent truth exists, science has no privileged means of accessing it.) Our definition of natural valuation is in effect a proxy for the scientific community's opinion on the truth of a claim. Accordingly, Theorem 2 says that the provisional proof procedure neither under-generates nor over-generates defeasibly true claims, provided all grounded arguments for claims are eventually asserted.

## 5 EXAMPLE

To illustrate these ideas we present a simple and hypothetical example of an Agora debate. In a real debate, participants would be free to introduce supporting evidence and modes of inference at any time. For reasons of space, in this example we first list the statements and modes of inference to be asserted, labeled K1 through K4, and R1 through R3, respectively, about a chemical $\mathcal{X}$:

**K1:** $\mathcal{X}$ is produced by the human body naturally (i.e. it is endogenous).

**K2:** $\mathcal{X}$ is endogenous in rats.

**K3:** An endogenous chemical is not carcinogenic.

**K4:** Bioassay experiments applying $\mathcal{X}$ to rats result in significant carcinogenic effects.

**R1 (And Introduction):** Given a wff $\phi$ and a wff $\theta$, we may infer the wff $(\phi \wedge \theta)$.

**R2 (Modus Ponens):** Given a wff $\phi$ and the wff $(\phi \to \theta)$, we may infer the wff $\theta$.

**R3:** If a chemical is found to be carcinogenic in an animal species, then we may infer it to be carcinogenic in humans.

We now give an example of an Agora dialogue concerning the statement: $\mathcal{X}$ *is carcinogenic to humans*, which we denote by $\phi$. The moves are numbered M1, M2,..., in sequence, and for simplicity we assume the participants are using the claims dictionary of Example 1, abbreviated to $\{Cert, Conf, Prob, Plaus, Supp, Open\}$, and the inference dictionary $\mathcal{D}_I = \{Val, Inval\}$. Before any discourse move is made, Nature's modality for this claim is $d_{\phi,N} = Open$, as is its modality for $\neg \phi$. Ignoring claims about any other chemicals, we thus have at commencement that $\mathcal{CS}(\mathcal{P}_N) = \{(\phi, Open), (\neg\phi, Open)\}$. Through the dialogue, we show the contents of Nature's commitment store as it changes, in steps numbered NCS0, NCS1,...

**NCS0:** $\mathcal{CS}(\mathcal{P}_N) = \{(\phi, Open), (\neg\phi, Open)\}$.

**M1:** $assert(\mathcal{P}_1 : (\phi, Conf))$.

**M2:** $query(\mathcal{P}_2 : assert(\mathcal{P}_1 : (\phi, Conf)))$.

**M3:** $show\_arg(\mathcal{P}_1 : (K4, R3, \phi, (Conf, Val, Conf)))$.

**NCS1:** $\mathcal{CS}(\mathcal{P}_N) = \{(\phi, Conf), (\neg\phi, Open)\}$.

**M4:** $contest(\mathcal{P}_2 : assert(\mathcal{P}_1 : (\phi, Conf)))$.

**M5:** $query(\mathcal{P}_3 :contest(\mathcal{P}_2 :assert(\mathcal{P}_1 : (\phi, Conf))))$

**M6:** $propose(\mathcal{P}_2 : (\neg\phi, Plaus))$.

**M7:** $query(\mathcal{P}_1 :propose(\mathcal{P}_2 : (\neg\phi, Plaus)))$.

**M8:** $show\_arg(\mathcal{P}_2 : ((K1, K3), R2, \neg\phi,$
  $(Conf, Prob, Val, Plaus)))$.

**NCS2:** $\mathcal{CS}(\mathcal{P}_N) = \{(\phi, Plaus), (\neg\phi, Plaus)\}$.

**M9:** $contest\_ground(\mathcal{P}_4 :$
  $show\_arg(\mathcal{P}_2 : ((K1, K3), R2, \neg\phi,$
  $(Conf, Prob, Val, Plaus)) : (K3, Prob)))$.

**M10:** $show\_arg(\mathcal{P}_4 : ((K2, K4), R1, \neg K3,$
  $(Conf, Conf, Val, Conf)) )$

**NCS3:** $\mathcal{CS}(\mathcal{P}_N) = \{(\phi, Plaus), (\neg\phi, Plaus)\}$.

Observe that Participant $\mathcal{P}_4$ in Move **M10**, by providing an argument for $\neg K3$, undercuts the argument presented for $\phi$ by Participant $\mathcal{P}_2$ in Move **M8**. We can also observe the changes in the Natural Valuation of $\phi$ through the course of this debate. At the start, we have $v_N(\phi) = 0$, which changes to $v_N(\phi) = 1$ after Move **M3**, since then $d_{\theta,N} = Conf$. However, after Move **M8**, $d_{\theta,N} = Plaus$, so once again $v_N(\phi) = 0$.

## 6 DISCUSSION

Characterization of scientific discourse as dialectical argumentation is not new. Rescher (1977) claims to have been the first to propose a dialectical framework for modeling the progress of scientific inquiry, and Pera's (1994) work is also a dialectical approach to science. Among argumentation theorists, Freeman (1991) also discusses scientific discourse in his study of argument structure. Both Carlson (1983) and Walton and Krabbe (1995) aim to model generic dialogues, but their focus is (respectively) on question-and-answer and persuasion dialogues. In addition, neither formalism explicitly permits degrees of support for commitments to be expressed, which our formalism does.

None of these works appears intended for encoding in intelligent systems. Within AI, intelligent systems for scientific domains have used argumentation for some time (e.g. Fox, Krause, & Ambler 1992). However, these applications have typically involved monolectical rather than dialectical argumentation. More recently, Haggith (1996) developed a dialectical argumentation formalism and applied the resulting system to a carcinogenicity debate. However, the primary focus of her work was on knowledge representation in generic domains of conflict, and so her formalism is not grounded in an explicit philosophy of science. The work of Amgoud, Maudet, & Parsons (2000) is closest in approach to that presented here (and we have drawn upon their formalism), but it is focused on negotiation dialogues, again in a generic context. Their formalism only permits two participants, although this would be relatively easy to amend. As with Haggith's system, their formalism does not permit debate over the rules of inference used. Recent legal argumentation systems, such as those of Verheij (1999), do permit this.

Our formal definition of the Risk Agora enables contestation and defeasibility of scientific claims. Our system therefore operationalizes the two normative principles of conduct for scientific discourses presented in Section 2.1. We are currently exploring a number of refinements to the Agora. Firstly, Rehg (1997) has demonstrated the rationality of incorporation of rhetorical devices (such as epideictic speech and appeals to emotions) in dialectical argument and decision-making, and we seek a means to incorporate such devices in the Agora. This would not be novel: the argumentation system of Reed (1998), for example, allows for the modeling of rhetorical devices, although in a monolectical context. Secondly, using the Agora in a deliberative context would require incorporation of values for the projected consequences and the development of an appropriate qualitative decision-theory, as in (Fox & Parsons 1998; Parsons & Green 1999).

We believe the Risk Agora has a number of potential benefits. Firstly, by articulating precisely the arguments used to assert carcinogenicity, gaps in knowledge and weaknesses in arguments can be identified more readily. Such identification could be used to prioritize bio-medical research efforts for the particular chemical. Secondly, by exploring the logical consequences of claims, the Risk Agora can serve a social maieutic function, making explicit knowledge which may only be latent. Thirdly, once instantiated with the details of a particular debate, the system could be used for self-education by others outside the scientific community concerned. Indeed, it could potentially form the basis for the making of regulatory or societal decisions on the issues in question (e.g. *Should the chemical be banned?*), and thereby give practical effect to notions of deliberative democracy (McBurney & Parsons 2000a; 2000 In press). Finally, with argumentation increasingly being used in the design of multi-agent systems (Parsons, Sierra, & Jennings 1998), the formalism presented here could readily be adapted for deliberative dialogues between independent software agents.

### Acknowledgments

This work was partly funded by the British EPSRC under grant GR/L84117 and a PhD studentship. We also thank the anonymous reviewers for their comments.

UNCERTAINTY IN ARTIFICIAL INTELLIGENCE PROCEEDINGS 2000                                                                                     379*tive Ethics Controversy*. Cambridge, MA, USA: MIT Press. 151–190. *(Published in German 1978)*.

Amgoud, L.; Maudet, N.; and Parsons, S. 2000. Modelling dialogues using argumentation. In Durfee, E., ed., *Proceedings of the 4th International Conference on Multi-Agent Systems (ICMAS-2000)*. Boston, MA, USA: IEEE.

Carlson, L. 1983. *Dialogue Games: An Approach to Discourse Analysis*. Dordrecht, The Netherlands: D. Reidel.

E.P.A. U.S.A. 1986. Guidelines for carcinogen risk assessment. *U.S. Federal Register* 51:33991–34003.

Feyerabend, P. 1993. *Against Method*. London, UK: Verso, third edition. First edition published 1971.

Fox, J., and Parsons, S. 1998. Arguing about beliefs and actions. In Hunter, A., and Parsons, S., eds., *Applications of Uncertainty Formalisms*, LNAI 1455. Berlin, Germany: Springer Verlag. 266–302.

Fox, J.; Krause, P.; and Ambler, S. 1992. Arguments, contradictions and practical reasoning. *Proceedings of ECAI 1992, Vienna, Austria*.

Freeman, J. B. 1991. *Dialectics and the Macrostructure of Arguments: A Theory of Argument Structure*. Berlin, Germany: Foris.

Graham, J. D.; Green, L. C.; and Roberts, M. J. 1988. *In Search of Safety: Chemicals and Cancer Risk*. Cambridge, MA, USA: Harvard University Press.

Grice, H. P. 1975. Logic and conversation. In Cole, P., and Morgan, J. L., eds., *Syntax and Semantics III: Speech Acts*. New York City, NY, USA: Academic Press. 41–58.

Habermas, J. 1991. *Moral Consciousness and Communicative Action*. Cambridge, MA, USA: MIT Press. *(Published in German 1983)*.

Haggith, M. 1996. *A Meta-level Argumentation Framework for Representing and Reasoning about Disagreement*. Ph.D. Dissertation, University of Edinburgh, UK.

Hamblin, C. L. 1971. Mathematical models of dialogue. *Theoria* 37:130–155.

Krause, P.; Fox, J.; Judson, P.; and Patel, M. 1998. Qualitative risk assessment fulfils a need. In Hunter, A., and Parsons, S., eds., *Applications of Uncertainty Formalisms*, LNAI 1455. Berlin, Germany: Springer Verlag. 138–156.

Lakatos, I., and Feyerabend, P. 1999. *For and Against Method*. Chicago, IL, USA: University of Chicago Press.

McBurney, P., and Parsons, S. 1999. Truth or consequences: using argumentation to reason about risk. In *British Psychological Society Symposium on Practical Reasoning*. London, UK: BPS.

McBurney, P., and Parsons, S. 2000a. Intelligent systems to support deliberative democracy in environmental regulation. In Peterson, D., ed., *AISB Symposium on AI and Legal Reasoning*. Birmingham, UK: AISB.

McBurney, P., and Parsons, S. 2000b. Modeling scientific discourse. In Pearce, D., ed., *Proceedings of the Workshop on Scientific Reasoning in AI and Philosophy of Science, 14th European Conference on Artificial Intelligence (ECAI-2000)*. Berlin, Germany: ECAI.

McBurney, P., and Parsons, S. 2000c. Tenacious tortoises: a formalism for argument over rules of inference. Technical report, Department of Computer Science, University of Liverpool, UK.

McBurney, P., and Parsons, S. 2000 (In press). Risk Agoras: Using dialectical argumentation to debate risk. *Risk Management Journal*.

Naess, A. 1966. *Communication and Argument: Elements of Applied Semantics*. London, UK: Allen and Unwin. *(Published in Norwegian 1947)*.

Parsons, S., and Green, S. 1999. Argumentation and qualitative decision making. In Hunter, A., and Parsons, S., eds., *The 5th European Conference on Symbolic and Quantitative Approaches to Reasoning and Uncertainty (ECSQARU99)*, LNAI 1638. Berlin, Germany: Springer Verlag. 328–339.

Parsons, S.; Sierra, C.; and Jennings, N. R. 1998. Agents that reason and negotiate by arguing. *Journal of Logic and Computation* 8(3):261—292.

Pera, M. 1994. *The Discourses of Science*. Chicago, IL, USA: University of Chicago Press.

Reed, C. 1998. *Generating Arguments in Natural Language*. Ph.D. Dissertation, University College, University of London, London, UK.

Rehg, W. 1997. Reason and rhetoric in Habermas's Theory of Argumentation. In Jost, W., and Hyde, M. J., eds., *Rhetoric and Hermeneutics in Our Time: A Reader*. New Haven, CN, USA: Yale University Press. 358–377.

Rescher, N. 1977. *Dialectics: A Controversy-Oriented Approach to the Theory of Knowledge*. Albany, NY, USA: State University of New York Press.

Toulmin, S. E. 1958. *The Uses of Argument*. Cambridge, UK: Cambridge University Press.

Verheij, B. 1999. Automated argument assistance for lawyers. In *Proceedings of the Seventh International Conference on Artificial Intelligence and Law, Oslo, Norway*, 43–52. New York City, NY, USA: ACM.

Walton, D. N., and Krabbe, E. C. W. 1995. *Commitment in Dialogue: Basic Concepts of Interpersonal Reasoning*. Albany, NY, USA: State University of New York Press.